\documentclass[11pt,a4paper]{article}
\usepackage[hyperref]{emnlp2018}
\usepackage{times}
\usepackage{latexsym}

\usepackage{url}

\DeclareTextFontCommand{\emph}{\em}
\usepackage{float}
\usepackage[utf8]{inputenc}
\usepackage[T1]{fontenc}
\usepackage{placeins}
\usepackage[subrefformat=parens]{subcaption}
\usepackage{amsmath, amssymb}
\usepackage{type1cm}
\usepackage{enumitem}
\usepackage{multirow}
\usepackage{balance}

\newcommand{\softmax}{\mathop{\rm softmax}\limits}
\global\long\def\T#1{#1^{\top}}

\DeclareMathAlphabet{\mathpzc}{OT1}{pzc}{m}{it}
\usepackage{mathtools}
\usepackage{bm}
\usepackage{algorithm}
\usepackage{algorithmicx}
\usepackage{algpseudocode}
\algtext*{EndWhile}
\algtext*{EndIf}
\algtext*{EndFor}
\usepackage{graphicx}

\aclfinalcopy 
\def\aclpaperid{***} 

\title{End-to-End Argument Mining for Discussion Threads\\ Based on Parallel Constrained Pointer Architecture}

\author{Gaku Morio \and Katsuhide Fujita \\
Tokyo University of Agriculture and Technology\\
2-24-16, Koganei, Tokyo, Japan\\
{\tt morio@katfuji.lab.tuat.ac.jp, katfuji@cc.tuat.ac.jp}
}

\date{}

\begin{document}

\setlength{\abovedisplayskip}{-1pt} 
\setlength{\belowdisplayskip}{1pt} 

\twocolumn[ 
  {\bf End-to-End Argument Mining for Discussion Threads Based on Parallel Constrained Pointer Architecture}\\
  Gaku Morio and Katsuhide Fujita\\
  Tokyo University of Agriculture and Technology, Japan.
  \\
  \par
  This paper is a pre-print version of the paper accepted for publication of 5th Workshop on Argument Mining (ArgMining) at 2018 Conference on Empirical Methods in Natural Language Processing (EMNLP 2018).
  Please use the following bib format until the the final version is available on the ACL Anthology: {\href{http://aclweb.org/anthology/}{http://aclweb.org/anthology/}}.
  \\
  \\
  @InProceedings\{morio2018.argmining,\\
  \, title = \{End-to-End Argument Mining for Discussion Threads Based on Parallel Constrained Pointer Architecture\},\\
  \, author = \{Morio, Gaku and Fujita, Katsuhide\},\\
  \, booktitle = \{Proceedings of the 5th Workshop on Argument Mining\},\\
  \, year      = \{2018\},\\
  \, address   = \{Brussels, Belgium\},\\
  \, publisher = \{Association for Computational Linguistics\},\\
  \, pages = \{(to appear)\}\\
  \}
]

\clearpage

\maketitle

\begin{abstract}
Argument Mining (AM) is a relatively recent discipline, which concentrates on extracting claims or premises from discourses, and inferring their structures. However, many existing works do not consider micro-level AM studies on discussion threads sufficiently.
In this paper, we tackle AM for discussion threads. Our main contributions are follows: (1) A novel combination scheme focusing on micro-level inner- and inter- post schemes for a discussion thread. (2) Annotation of large-scale civic discussion threads with the scheme. (3) {\it Parallel constrained pointer architecture} (PCPA), a novel end-to-end technique to discriminate sentence types, inner-post relations, and inter-post interactions simultaneously.\footnote{Available at:\\ \href{https://github.com/EdoFrank/EMNLP2018-ArgMining-Morio}{https://github.com/EdoFrank/EMNLP2018-ArgMining-Morio} including source codes.}
The experimental results demonstrate that our proposed model shows better accuracy in terms of relations extraction, in comparison to existing state-of-the-art models.
\end{abstract}

\section{Introduction}
Argument Mining (AM) is a discipline which concentrates on extracting claims or premises, and inferring their structures from a discourse. In \cite{Palau:2009,Stab2014,Peldszus:2013}, they construed an argument as the pairing of a single claim and a (possibly empty) set of premises, which justifies the claim.

\begin{figure*}[tb]
  \begin{center}
  \includegraphics[width=0.55\linewidth]{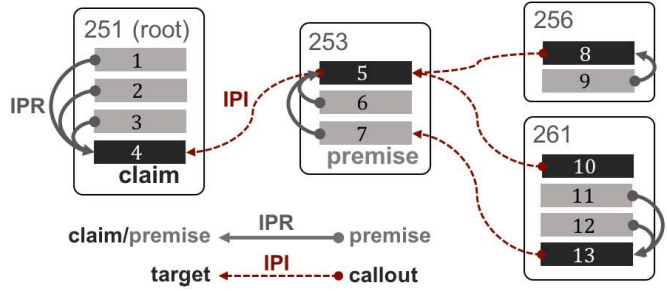}
  \end{center}
  \caption{Example of our scheme for a thread.}
  \label{fig:thread}
\end{figure*}

\par
Generally, identifying structures for argument components (i.e., premises and claims) is categorized as a micro-level approach, and among complete arguments as a macro-level approach. There are some micro-level approaches \cite{Palau:2009,Stab2014,stab2017}, however, few AM studies aggressively consider a scheme of micro-level reply-to interactions in a thread. Though \newcite{hidey2017} provided a micro-level thread structured dataset, they considered an entire thread as a discourse. Thus, they allowed a premise that links to a claim in another post, while a post should be considered as a stand-alone discourse because a writer for each post is different. Also, we need to consider post-to-post interactions with the stand-alone assumption as a backdrop. Moreover, the dataset of \cite{hidey2017} with only 78 threads is too small to apply state-of-the-art neural discrimination models.

\par
In addition to the shortage of micro-level anotations for discussion threads, no empirical study on end-to-end discrimination models which tackle discussion threads exist, to the best of our knowledge.

\par
Motivated by the weaknesses above, this paper commits to the empirical study for discussion threads. Our main three contributions are as follows: (1) A novel combination scheme to apply AM to discussion threads. We introduce {\it inner-post} and {\it inter-post} schemes in combination. This combination enables us to discriminate arguments per post, rather than per thread as in \cite{hidey2017}. In the former scheme, a post is assumed as a stand-alone discourse and a micro-level annotation is provided. In the second scheme, we introduce inter-post micro-level interactions. The introduction of the interactions allows us to capture informative argumentative relations between posts. (2) Large-scale online civic discussions are annotated by the proposed scheme. Specifically, we provide two phase annotation, and evaluate inter-annotator agreements. (3) A {\bf parallel constrained pointer architecture} (PCPA) is proposed, which is a novel end-to-end neural model. The model can discriminate types of sentences (e.g., claim or premise), inner-post relations and inter-post interactions, simultaneously.
In particular, our PCPA achieved a significant improvement on challenging relation extractions in comparison to the existing state-of-the-art models \cite{eger2017,Potash2017}. An advantage of our model is that the constraints of a thread structure are considered. The constraints make our architectures effective at learning and inferring, unlike existing pointer models. 

\par
While our dataset of discussion threads will make further advances in AM, the proposed PCPA will make end-to-end AM studies going forward.

\section{Related Works} \label{RelatedWorks}
\newcite{stab2017} argue that the task of AM is divided into the following three subtasks:
\begin{itemize}
\small
\setlength{\parskip}{0cm} 
\setlength{\itemsep}{0cm} 
\item {\bf Component identification} focuses on separation of argumentative and non-argumentative text units and identification of argument component boundaries.
\item {\bf Component classification} addresses the function of argument components. It aims at classifying argument components into different types, such as claims and premises.
\item {\bf Structure identification} focuses on linking arguments or argument components. Its objective is to recognize different types of argumentative relations, such as support or attack relations.
\end{itemize}

\noindent
The structure identification can also be divided to macro- and micro-level approaches. The macro-level approach as in \cite{boltuvzic-vsnajder:2014, gosh2014, Murakami:2010} addresses relations between complete arguments and ignores the micro-structure of arguments \cite{stab2017}. In \cite{gosh2014}, the authors introduced a scheme to represent relations between two posts by {\it target} and {\it callout}; however, their study discards micro-level structures in arguments because of their macro-level annotation. The micro-level approach as in \cite{Palau:2009,Stab2014,stab2017} focuses on the relations between argument components. In \cite{Palau:2009}, arguments are considered as trees. In \cite{stab2017}, the authors also represented relations of argument components in essays as tree structures. However, they addressed discourses of a single writer (i.e., an essay writer) rather than multiple authors in a discussion thread. Therefore, we can't simply apply their scheme to our study.

Recently, the advances of automatic detection of argument structures have been seen in the discipline of AM. Some recent papers \cite{Lippi:2015,ecklekohler-kluge-gurevych:2015} propose argument component identification to extract argumentative components in the entire discourse. These works \cite{Persing2016,eger2017,Potash2017} showed link extraction task to find argumentative relations between argument components.
\par
End-to-end discrimination models are also highlighted in AM. The reason is low error propagation compared with the other ends (pipeline). The pipeline models have to discriminate argument component identification and link extraction subtasks independently, and thus cause the error propagation \cite{eger2017}. The authors propose manners to apply multi-task learning \cite{Sogaard2016, martinezalonso2017} and LSTM-ER \cite{Miwa2016} to the end-to-end AM. Another end-to-end work for AM, \newcite{Potash2017} argues that Pointer Networks \cite{Vinyals2015, Katiyar2017} which incorporate a sequence-to-sequence model in their classifier is a state-of-the-art model for argument component type prediction and link extraction tasks.

\section{Argument Mining for Discussion Thread} \label{ArgumentMiningforDiscussionThread}

\subsection{Scheme}
In this work, we present a novel scheme combining {\it inner-post} scheme of a stand-alone post with {\it inter-post} scheme that considers a reply-to argumentative relation. In the inner-post scheme (e.g., claim/premise types and inner-post relations), "one-claim" approach from \cite{stab2017} is adopted. In the inter-post scheme, the micro-level interaction in the spirit of \cite{gosh2014} is employed. The definitions of inner-post relation and inter-post interaction are follows:

\begin{itemize}[leftmargin=*]
\small
\setlength{\parskip}{0cm} 
\setlength{\itemsep}{0cm} 
\item {\bf Inner-post relation (IPR)} is a directed argumentative relation in a post. Each IPR:$(target \leftarrow source)$ indicates that the $source$ component is either a justification for or a refutation of the $target$ component. Thus, a $source$ should be a premise, and each premise has a single outgoing link to another premise or claim \cite{eger2017}.
\item {\bf Target} is a head of IPI that has been called out by a subsequent claim in another post that replies to the post of the target.
\item {\bf Callout} is a tail of IPI that refers back to a prior target. In addition to referring back to the target, a callout must be a claim.\footnote{To restrict a callout to a claim makes our problem more simple because the number of outgoing links from a claim becomes one at a maximum. Thus, we introduced the restriction.}
\item {\bf Inter-post interaction (IPI)} is the micro-level relationship of two posts: parent post and child post that replies to the parent post. A relation $(parent \leftarrow child)$ represents the $child$ is a {\bf callout} and $parent$ is a {\bf target}.
\end{itemize}

\noindent
Figure \ref{fig:thread} shows our combination scheme for a discussion thread.

\subsection{Dataset}
To develop a sufficient AM corpus for discussion threads, we have annotated an original large-scale online civic discussion \cite{morio2018b}. The civic discussion data is obtained by an online civic engagement on the {\it COLLAGREE} \cite{Ito2014,morio2018} including a thread structure. The discussion was held from the end of 2016 to the beginning of 2017, and co-hosted by the government of Nagoya City, Japan. The accumulated data includes 204 citizens, 399 threads, 1327 posts, 5559 sentences and 120241 tokens spelled in Japanese.\footnote{The average of the number of posts per thread is 3.33 (standard deviation is 3.29), the depth of threads is 1.09 (standard deviation is 1.19), the number of sentences per post is 4.19 (standard deviation is 3.33) and the number of words per sentence is 21.63 (standard deviation is 19.92).} To the best of our knowledge, this work is the first approach which annotates large-scale civic discussions for AM.\footnote{Recently, \newcite{Park2018} provide a similar dataset of civic engagement, while their dataset doesn't consider post-to-post relations sufficiently.}

\subsection{Annotation Design}
In \cite{Peldszus:2013}, the authors argue that the annotation task for AM contains the following three subtasks: (1) segmentation, (2) segment classification and (3) relationship identification. The segmentation requires extensive human resources, time, and cost. Therefore, we apply a rule-based technique for the segmentation. 
Then, we consider each sentence as an argument component candidate (ACC). For classifying the argument component, the ACC types (claim, premise or non-argumentative (NonArg)) for each ACC are annotated. Finally, the relationship identification needs to annotate IPRs and IPIs.

\par
Using multiple processes for multiple annotation subtasks is common \cite{Meyers2010,Stab2014,stab2017}. To annotate our data, we provide two phases. 
In the first phase, we concentrate on annotating ACC type and IPR, and create a temporal gold standard. In the second phase, IPI is annotated using the temporal gold standard.

\par
We employed a majority vote to create the gold standard. All three annotators independently annotated in this work. 
The procedure of the first phase for compiling the temporal gold standard is as follows.

\begin{enumerate}[leftmargin=*]
\small
\setlength{\parskip}{0cm} 
\setlength{\itemsep}{0cm} 
\item [A1:] Each ACC type is decided on a majority vote. When the ACC type of the sentence cannot be decided by majority vote, NonArg is assigned to them.
\item [A2:] Each IPR (link existence) is decided on a majority vote.
\item [A3:] Merging the results from A1 and A2, and obtaining trees where root is a claim. Thus, we have trees to the number of claims in a post.
\item [A4:] Eliminating premise tags that do not belong to any trees, assigning them to NonArg, and eliminating their IPR.
\end{enumerate}



\subsection{Annotation Result}
Inter-annotator agreement for ACC type, IPR and IPI annotations are calculated using Fleiss's $\kappa$ \cite{fleiss1971}. First, we attempt to evaluate the agreement of the first phase annotations, however, the $\kappa$ of IPR is relatively low: $0.420$. 
The annotators are less likely to agree on serial arguments \cite{stab2017} like $(premise \leftarrow premise)$ relations.\footnote{Unlike with Persuasive Essays \cite{stab2017}, citizen's documents for civic discussions are seldom well-structured. Thus, we don't see the point in providing a more complex scheme (i.e., allowing $(premise \leftarrow premise)$ relations).} Therefore, we introduce an initial process A0, transforming $(premise1 \leftarrow premise2)$ into $(\text{\it root claim of premise2} \leftarrow premise2)$, before A1.\footnote{For example, two IPRs $\{(claim1 \leftarrow premise1), (premise1 \leftarrow premise2)\}$ are transformed to $\{(claim1 \leftarrow premise1), (claim1 \leftarrow premise2)\}$.}

\begin{table}[tb]
\centering
\small
\begin{tabular}{|l|l|c|c|}
\hline
\multicolumn{1}{|c|}{Corpus}       & \multicolumn{1}{c|}{Type} & Size & $\kappa$ \\ \hline
\multirow{5}{*}{COLLAGREE}         & Claim                     & 1449 & .531     \\
                                   & Premise                   & 2762 & .554     \\
                                   & NonArg                    & 1348 & .529     \\
                                   & IPR w/ A0       & 2762 & .466     \\
                                   & IPI    & 745  & .430     \\ \hline
\multirow{3}{*}{Persuasive Essays} & Claim                     & 1506 & .635     \\
                                   & Premise                   & 3832 & .833     \\
                                   & Inner-essay rel      & 3832 & .708-.737        \\ \hline
\end{tabular}
\caption{Inter-annotator agreement scores for the two corpora.}
\label{tb:kappa}
\end{table}

\begin{figure*}[tb]
  \begin{center}
  \includegraphics[width=320pt]{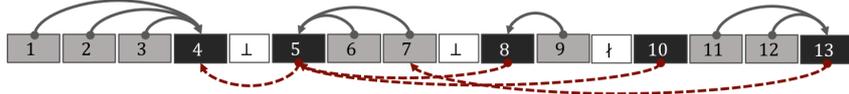}
  \end{center}
  \caption{Sequence representation from Figure \ref{fig:thread}. $\bot \text{ and } \mid \hspace{-0.75em}/$ denotes a separator representation of thread depth and posts.}
  \label{fig:encoding}
\end{figure*}

\par
Table \ref{tb:kappa} summarizes the number of each type of relation and inter-annotator agreement.\footnote{Outgoing IPI links are composed of 574 claims, 109 premises, and 62 NonArgs. Considering that a callout should be a claim, the $(claim \leftarrow claim)$ interaction accounts for 77\% of the total. The results indicate that IPIs are pretty argumentative. In addition, we annotated support/attack relations \cite{Cocarascu2017}. The results show support accounts for 86\% and attacks for 7\% of the total IPIs.} For comparison, we also mention the annotation results of Persuasive Essays \cite{stab2017}. Unlike the essay dataset, our datasets contain badly-structured writings, resulting in low agreement. However, classification tasks can be applied as \cite{Landis77} refers to the $\kappa$ value from 0.41 to 0.61 as "moderate agreement". Moreover, the agreement of IPR is improved by providing the process A0.

\section{Discriminating ACC Type, Inner-Post Relation and Inter-Post Interaction} \label{Discriminating}
This section describes the study on our end-to-end discrimination model, which identifies ACC type, IPR and IPI for our annotated dataset.

\subsection{Thread Representation as a Sequence} \label{ThreadRepresentation}

If the thread itself contains flow of its argument, only the thread itself is considered as the desirable input for a discrimination model. Thus, we describe a way of representing a thread with an input sequence.

\par
In this work, we extend the sequence representation of \cite{eger2017, Potash2017}. The creation of thread representation as an input sequence consists of the following two steps. First, we assume each element of the input sequence for recurrent neural network is a sentence representation, rather than a word representation. Second, we sort the sentence representations by the thread depth order. In addition, for each thread depth, we in turn order them according to the timestamp of their post, and insert separator representations. The first one makes it possible to input a short sequence to LSTM units \cite{Hochreiter:1997}. The second makes a classifier easy to discriminate considering the hierarchy of a thread and reply relations. Figure \ref{fig:encoding} shows an example of a thread representation as sequence.

\subsection{Parallel Constrained Pointer Architecture}
One of the main technical contributions of our approach is to provide a discrimination model that classifies ACC type, IPR and IPI simultaneously via end-to-end learning. A Pointer Network (PN) for end-to-end AM achieves state-of-the-art results \cite{Potash2017}, which leads to applying a PN based technique to our scheme. Unfortunately, the naive PN did not achieve the result expected (the quantitative results are shown in Section \ref{Experiments}), because the simple PN is unable to constrain its search space for thread structures. For instance, an inner-post relation classifier could discriminate with no need to search out of its post, or an inter-post interaction classifier could classify with no need to search out of the parent post and child post. Therefore, we propose a novel neural model named {\it parallel constrained pointer architecture} (PCPA). PCPA provides two parallel pointer architectures: IPR and IPI discrimination architectures that adopt the apparent constrains of threads.

\begin{figure*}[tb]
  \begin{center}
  \includegraphics[width=0.8\linewidth]{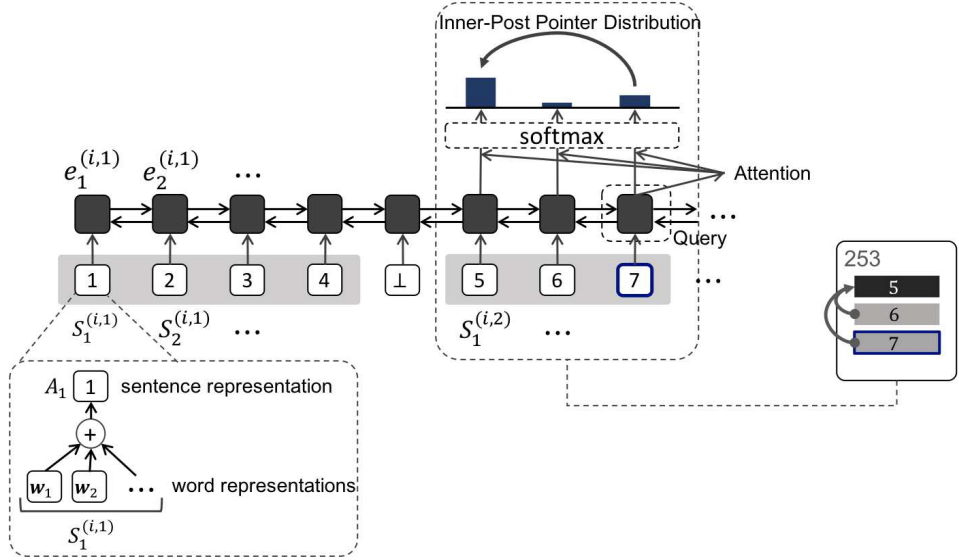}
  \end{center}
  \caption{Example of the constrained pointer architecture of inner-post relation (IPR) identification, discriminating the IPR target from the ACC "7".}
  \label{fig:network_inner}
\end{figure*}

\subsubsection*{Sentence Representation as Input}
First, we introduce the input representation. Given $N$ threads $(T_1,\dots,T_N)$, we denote $T_i$'s posts which are sorted in thread depth order, and then timestamp order as described in Section \ref{ThreadRepresentation} as $(P^{(i)}_1,\dots,P^{(i)}_{N_i})$, where $N_i$ represents the number of posts in $T_i$. In addition to the thread and post representations, write $(S^{(i,j)}_1,\dots,S^{(i,j)}_{N_{i,j}})$ for sentences in post $P^{(i)}_j$, where $N_{i,j}$ represents the number of sentences in $P^{(i)}_j$. Note that separator representations are not considered in the notation.

\par
Then, $\bm{w}_n$ is given initially, an embedding vector of $n$th word in a sentence $S^{(i,j)}_k$, a sentence representation for an input of LSTM is represented as: $A_k = \sum_n \bm{w}_n$, where $\bm{w}_n$ is gained from bag-of-words (BoW) or word embeddings \cite{Mikolov2013, Pennington2014, stab2017}. 
In our study, we employed BoW and a fully connected layer with a trainable parameter to learn word embeddings.
Subsequently, we provide Bidirectional LSTM (BiLSTM) \cite{Graves05} because PN requires encoding steps. At each time step of the encoder BiLSTM, PCPA considers a representation of an ACC. Thus, the hidden representation $e_i$ of BiLSTM becomes the concatenation of forward and backward hidden representations. To simplify the explanation, we denote the hidden representations of $(S^{(i,j)}_1,\dots,S^{(i,j)}_{N_{i,j}})$ as $(e^{(i,j)}_1,\dots,e^{(i,j)}_{N_{i,j}})$. For better understanding, we show notations in Figure \ref{fig:network_inner}.

\subsubsection*{Discriminating Inner-Post Relation}

\begin{figure*}[tb]
  \begin{center}
  \includegraphics[width=0.8\linewidth]{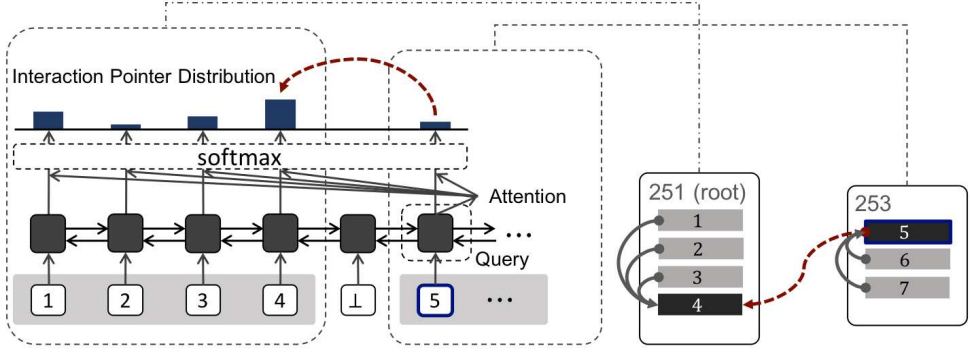}
  \end{center}
  \caption{Example of the constrained pointer architecture of inter-post interaction (IPI) identification, discriminating the IPI target that is called out from the ACC "5".}
  \label{fig:network_inter}
\end{figure*}

\par
The general PN of \cite{Potash2017} uses all hidden states $e_i$. Alternately, PCPA can limit the states to improve the accuracies, since each premise has a single outgoing link to another sentence in its post. Hence, we provide an approach to discriminate IPR using only inner-post hidden states of the BiLSTM. 
\par
Figure \ref{fig:network_inner} shows the example IPR discrimination in thread $T_i$; for example, we assume that the inner-post relation of the sentence written as "7" in the 3rd ACC of post $P^{(i)}_2$ is classified. The general PN needs to consider all $e_i$, therefore, the search space is large. On the other
hands, our proposed PCPA can consider $(e^{(i,2)}_1,e^{(i,2)}_2,e^{(i,2)}_3)$, which needs to use the hidden states of its post only. Therefore, our constrained architecture can reduce the search space significantly.

\par
In general, given $W_1$, $W_2$, and $v_1$, parameters of attention model \cite{Luong2015} for PN,

\begin{align}
\label{eq:uijkl}
u^{(i,j,k)}_l = \T{v_1} \tanh \left( W_1 e^{(i,j)}_l + W_2 e^{(i,j)}_k \right)
\end{align}

\noindent
represents a degree that $k$th ACC in post $P^{(i)}_j$ has an outgoing link to $l$th ACC. Moreover, we can assume $e^{(i,j)}_k$ as a query vector. Supposing the ACC has no outgoing link, we can consider the ACC learned to point to itself.
Although equation (\ref{eq:uijkl}) is real-value, a distribution over the IPR input is considered by taking $\softmax$ function, i.e.,
\begin{align}
\label{eq:softmax_inner}
p ( y^{ipr}_k \mid P^{(i)}_j ) = \softmax ( u^{(i,j,k)} )
\end{align}
representing the probability that $k$th ACC in post $P^{(i)}_j$ has an outgoing link to $l$th ACC in $P^{(i)}_j$. Therefore, the objective for IPR in thread $T_i$ is calculated by taking the sum of log-likelihoods for all posts:
\begin{align}
\label{eq:likelihood_inner}
L^{ipr}_i = \sum^{N_i}_{j=1} \sum^{N_{i,j}}_{k=1} \log p ( y^{ipr}_k \mid P^{(i)}_j)
\end{align}

\subsubsection*{Discriminating Inter-Post Interaction}

As the definition of target and callout in our scheme, IPI exists between a parent post and child post that replies to the parent. Thus, PCPA can discriminate IPI with no need to use all of the hidden representations of the LSTM. In other words, it can discriminate IPI without searching outside of the two posts.

\par
Hence, we design an output layer that requires only a set of reply pairs in thread $T_i$. Specifically, we assume that $R^{(i)}=\{ (j_1, j_2),\cdots \}$ where $j_1 \neq j_2 \land j_1 < j_2$ for a set of parent-child pairs in thread $T_i$. Supposing $j_1$ is the index of a parent post and $j_2$ represents the index of the child post that replies to the $j_1$. Note that when thread $T_i$ does not have any reply pairs, $R^{(i)}=\varnothing$. Considering the above, a technique that is similar to the IPR's technique is introduced.

\par
Figure \ref{fig:network_inter} shows the example IPI discrimination in thread $T_i$; supposing that we are going to discriminate a target that is called out from ACC "5" in the figure. In this case, the search space is limited by the parent post $(e^{(i,1)}_1,\dots,e^{(i,1)}_4)$. Moreover, we add an element $e^{(i,2)}_1$ so that a callout can point itself if there's no target in its parent post. The left four outputs in the "Interaction Pointer Distribution" indicate a discrete probabilistic distribution that the callout ACC "5" links to target sentences in its parent post, and an output on the far right represents a probability that the callout links to itself.

\par
The equation (\ref{eq:uijkl}) uses a query in the PN, so we in turn concentrate on using a query vector for the callout in IPI. Herein, we introduce an additional PN for IPI using new attention parameters, $W_3$,$W_4$ and $v_2$, as:

\begin{align}
\label{eq:qijkl}
q^{(i,j,k)}_l = \T{v_2} \tanh \left( W_3 e^{(i,j)}_l + W_4 e^{(i,j)}_k \right) 
\end{align}

\noindent
where $e^{(i,j)}_k$ is the query from the callout. Supposing that the reply pair is $(j_1, j_2)$, a target of $k$th ACC of the child post $P^{(i)}_{j_2}$ is searched. The expanded vector $\left[q^{(i,j_1,k)};q^{(i,j_2,k)}_k\right]$ is obtained by concatenating the attention vectors $q^{(i,j_1,k)}$ from the parent post and a vector $q^{(i,j_2,k)}_k$ from the callout. This expansion process is the same as the process of \cite{Merity2016}. Finally, given all reply pairs of thread $T_i$, the log-likelihood is calculated as follows:

\begin{align}
p ( y^{ipi}_k \mid P^{(i)}_{j_1}, P^{(i)}_{j_2} ) = \softmax ( [q^{(i,j_1,k)};q^{(i,j_2,k)}_k] ) \nonumber
\end{align}

\begin{align}
\label{eq:likelihood_inter}
L^{ipi}_i = \sum_{\left( j_1,{j_2} \right) \in R^{(i)}} \sum^{N_{i,j_2}}_{k=1} \log p ( y^{ipi}_k \mid P^{(i)}_{j_1}, P^{(i)}_{j_2} )
\end{align}

\subsubsection*{Discriminating ACC Type}
At each time step of the BiLSTM, the type classification task predicts whether it is claim, premise, or NonArg. The ACC type of sentence $S^{(i,j)}_k$ can be classified by taking $\softmax$ of $z^{(i,j)}_k = W_{type} e^{(i,j)}_k + b_{type}$, where $W_{type}$ and $b_{type}$ are parameters. An objective for the type classifier can also be described by taking the sum of log-likelihoods for all posts as:

\begin{align}
\label{eq:likelihood_type}
&p ( y^{type}_k \mid P^{(i)}_j ) = \softmax ( z^{(i,j)}_k ) \nonumber \\
&L^{type}_i = \sum^{N_i}_{j=1} \sum^{N_{i,j}}_{k=1} \log p ( y^{type}_k \mid P^{(i)}_j )
\end{align}

\subsubsection*{Joint Learning}
Combining objectives of IPR (equation (\ref{eq:likelihood_inner})), IPI (equation (\ref{eq:likelihood_inter})) and the ACC type (equation (\ref{eq:likelihood_type})), the training objective of PCPA is shown as follows:

\begin{align}
\label{eq:loss}
Loss = & \frac{1}{N} \sum_{i} ( - \alpha L^{ipr}_i  - \beta L^{ipi}_i\nonumber \\ 
       &- (1 - \alpha - \beta ) L^{type}_i )
\end{align}

\noindent
where $\alpha$ and $\beta$ are hyperparameters which adjust the weight of tasks in our cost function. Note that $\alpha,\beta \in[0,1] \land \alpha + \beta < 1$.

\section{Experiments} \label{Experiments}
\subsection{Experimental Settings}
\begin{table*}[tb]
\footnotesize
\centering
\begin{tabular}{llccccccc}
\hline
\multicolumn{1}{|c|}{\multirow{3}{*}{\begin{tabular}[c]{@{}c@{}}Model\\ type\end{tabular}}}                       & \multicolumn{1}{c|}{\multirow{3}{*}{Model name}} & \multicolumn{3}{c|}{Type classification}                                                                                                                                                                                                                                      & \multicolumn{4}{c|}{Link extraction}                                                                                                                                                                                                                                                                                                                                            \\ \cline{3-9} 
\multicolumn{1}{|c|}{}                                                                                            & \multicolumn{1}{c|}{}                            & \multicolumn{1}{c|}{\multirow{2}{*}{\begin{tabular}[c]{@{}c@{}}Claim\\ F1\end{tabular}}} & \multicolumn{1}{c|}{\multirow{2}{*}{\begin{tabular}[c]{@{}c@{}}Premise\\ F1\end{tabular}}} & \multicolumn{1}{c|}{\multirow{2}{*}{\begin{tabular}[c]{@{}c@{}}NonArg\\ F1\end{tabular}}} & \multicolumn{1}{c|}{\multirow{2}{*}{\begin{tabular}[c]{@{}c@{}}IPR\\ Precision\end{tabular}}} & \multicolumn{1}{c|}{\multirow{2}{*}{\begin{tabular}[c]{@{}c@{}}IPR\\ F1\end{tabular}}} & \multicolumn{1}{c|}{\multirow{2}{*}{\begin{tabular}[c]{@{}c@{}}IPI\\ Precision\end{tabular}}} & \multicolumn{1}{c|}{\multirow{2}{*}{\begin{tabular}[c]{@{}c@{}}IPI\\ F1\end{tabular}}} \\
\multicolumn{1}{|c|}{}                                                                                            & \multicolumn{1}{c|}{}                            & \multicolumn{1}{c|}{}                                                                    & \multicolumn{1}{c|}{}                                                                      & \multicolumn{1}{c|}{}                                                                 & \multicolumn{1}{c|}{}                                                                         & \multicolumn{1}{c|}{}                                                                  & \multicolumn{1}{c|}{}                                                                         & \multicolumn{1}{c|}{}                                                                  \\ \hline
\multicolumn{1}{|l|}{\multirow{5}{*}{\begin{tabular}[c]{@{}l@{}}Joint\\ learning\end{tabular}}}                   & \multicolumn{1}{l|}{Our Model}                   & \multicolumn{1}{c|}{58.5}                                                                & \multicolumn{1}{c|}{68.7}                                                                  & \multicolumn{1}{c|}{36.0}                                                             & \multicolumn{1}{c|}{33.8}                                                                     & \multicolumn{1}{c|}{*{\bf 40.8}}                                      & \multicolumn{1}{c|}{19.6}                                                                     & \multicolumn{1}{c|}{*{\bf 24.8}}                                      \\
\multicolumn{1}{|l|}{}                                                                                            & \multicolumn{1}{l|}{Our Model - Hyp}             & \multicolumn{1}{c|}{58.1}                                                                & \multicolumn{1}{c|}{{\bf 71.5}}                                           & \multicolumn{1}{c|}{{\bf 58.8}}                                      & \multicolumn{1}{c|}{*{\bf 45.8}}                                                                    & \multicolumn{1}{c|}{*{\bf 44.3}}                                      & \multicolumn{1}{c|}{*{\bf 30.4}}                                                                    & \multicolumn{1}{c|}{*{\bf 26.9}}                                      \\ \cline{2-9} 
\multicolumn{1}{|l|}{}                                                                                            & \multicolumn{1}{l|}{STagBLSTM}                   & \multicolumn{1}{c|}{54.2}                                                                & \multicolumn{1}{c|}{65.6}                                                                  & \multicolumn{1}{c|}{56.9}                                                             & \multicolumn{1}{c|}{14.3}                                                                     & \multicolumn{1}{c|}{14.9}                                                              & \multicolumn{1}{c|}{21.0}                                                                     & \multicolumn{1}{c|}{12.6}                                                              \\
\multicolumn{1}{|l|}{}                                                                                            & \multicolumn{1}{l|}{PN with Seq2Seq}                  & \multicolumn{1}{c|}{58.3}                                                                & \multicolumn{1}{c|}{70.8}                                                                  & \multicolumn{1}{c|}{48.6}                                                             & \multicolumn{1}{c|}{35.7}                                                                     & \multicolumn{1}{c|}{27.2}                                                              & \multicolumn{1}{c|}{13.0}                                                                     & \multicolumn{1}{c|}{19.4}                                                              \\
\multicolumn{1}{|l|}{}                                                                                            & \multicolumn{1}{l|}{PN without Seq2Seq}               & \multicolumn{1}{c|}{{\bf 60.1}}                                         & \multicolumn{1}{c|}{71.3}                                                                  & \multicolumn{1}{c|}{53.1}                                                             & \multicolumn{1}{c|}{36.6}                                                                     & \multicolumn{1}{c|}{35.0}                                                              & \multicolumn{1}{c|}{26.5}                                                                     & \multicolumn{1}{c|}{20.8}                                                              \\ \hline
                                                                                                                  &                                                  & \multicolumn{1}{l}{}                                                                     & \multicolumn{1}{l}{}                                                                       & \multicolumn{1}{l}{}                                                                  & \multicolumn{1}{l}{}                                                                          & \multicolumn{1}{l}{}                                                                   & \multicolumn{1}{l}{}                                                                          & \multicolumn{1}{l}{}                                                                   \\ \hline
\multicolumn{1}{|l|}{\multirow{3}{*}{\begin{tabular}[c]{@{}l@{}}Task\\ specific\end{tabular}}}                    & \multicolumn{1}{l|}{SVM - T}                     & \multicolumn{1}{c|}{53.3}                                                                & \multicolumn{1}{c|}{64.4}                                                                  & \multicolumn{1}{c|}{52.3}                                                             & \multicolumn{1}{c|}{13.8}                                                                     & \multicolumn{1}{c|}{22.4}                                                              & \multicolumn{1}{c|}{6.4}                                                                      & \multicolumn{1}{c|}{11.5}                                                              \\
\multicolumn{1}{|l|}{}                                                                                            & \multicolumn{1}{l|}{RF - T}                      & \multicolumn{1}{c|}{41.0}                                                                & \multicolumn{1}{c|}{66.8}                                                                  & \multicolumn{1}{c|}{38.3}                                                             & \multicolumn{1}{c|}{0}                                                                        & \multicolumn{1}{c|}{0}                                                                 & \multicolumn{1}{c|}{100}                                                                      & \multicolumn{1}{c|}{1.4}                                                               \\
\multicolumn{1}{|l|}{}                                                                                            & \multicolumn{1}{l|}{Simple - T}                  & \multicolumn{1}{c|}{41.1}                                                                & \multicolumn{1}{c|}{66.1}                                                                  & \multicolumn{1}{c|}{38.3}                                                             & \multicolumn{1}{c|}{0}                                                                        & \multicolumn{1}{c|}{0}                                                                 & \multicolumn{1}{c|}{0}                                                                        & \multicolumn{1}{c|}{0}                                                                 \\ \hline
                                                                                                                  &                                                  & \multicolumn{1}{l}{}                                                                     & \multicolumn{1}{l}{}                                                                       & \multicolumn{1}{l}{}                                                                  & \multicolumn{1}{l}{}                                                                          & \multicolumn{1}{l}{}                                                                   & \multicolumn{1}{l}{}                                                                          & \multicolumn{1}{l}{}                                                                   \\ \hline

\multicolumn{1}{|l|}{\multirow{4}{*}{\begin{tabular}[c]{@{}l@{}}Joint\\ learning\\ w/o\\ separator\end{tabular}}} & \multicolumn{1}{l|}{Our Model w/o separator}     & \multicolumn{1}{c|}{43.1}                                                                & \multicolumn{1}{c|}{66.3}                                                                  & \multicolumn{1}{c|}{29.6}                                                             & \multicolumn{1}{c|}{30.0}                                                                     & \multicolumn{1}{c|}{36.1}                                                              & \multicolumn{1}{c|}{9.9}                                                                      & \multicolumn{1}{c|}{13.7}                                                              \\
\multicolumn{1}{|l|}{}                                                                                            & \multicolumn{1}{l|}{STagBLSTM w/o separator}     & \multicolumn{1}{c|}{51.8}                                                                & \multicolumn{1}{c|}{66.1}                                                                  & \multicolumn{1}{c|}{55.2}                                                             & \multicolumn{1}{c|}{13.9}                                                                     & \multicolumn{1}{c|}{14.5}                                                              & \multicolumn{1}{c|}{16.1}                                                                     & \multicolumn{1}{c|}{10.8}                                                              \\
\multicolumn{1}{|l|}{}                                                                                            & \multicolumn{1}{l|}{PN with Seq2Seq w/o separator}    & \multicolumn{1}{c|}{40.7}                                                                & \multicolumn{1}{c|}{67.8}                                                                  & \multicolumn{1}{c|}{52.7}                                                             & \multicolumn{1}{c|}{30.4}                                                                     & \multicolumn{1}{c|}{23.2}                                                              & \multicolumn{1}{c|}{10.8}                                                                     & \multicolumn{1}{c|}{14.6}                                                              \\
\multicolumn{1}{|l|}{}                                                                                            & \multicolumn{1}{l|}{PN without Seq2Seq w/o separator} & \multicolumn{1}{c|}{43.4}                                                                & \multicolumn{1}{c|}{67.6}                                                                  & \multicolumn{1}{c|}{53.7}                                                             & \multicolumn{1}{c|}{29.5}                                                                     & \multicolumn{1}{c|}{21.1}                                                              & \multicolumn{1}{c|}{19.0}                                                                     & \multicolumn{1}{c|}{6.0}                                                               \\ \hline
\end{tabular}
\caption{{\bf Top}: Our models vs. joint baselines (\%). * indicates significant. at $p < 0.01$, two-sided Wilcoxon signed rank test \cite{Derryberry2010}, compared with each baseline. {\bf Middle}: Performances of task specific baselines. {\bf Bottom}: Performances of joint models w/o separator representations.}
\label{tb:f1_score}
\end{table*}

\subsubsection*{Evaluation Metric}
For the evaluation of ACC types, IPR and IPI discrimination, we adopt precision, recall and F1 scores. To obtain the precision and recall, we introduce a way to compute positive and negative cases by creating relations \cite{stab2017}, excluding self-pointers. \footnote{For example, supposing there is a post which contains three sentences, $(S_1, S_2, S_3)$, and two gold standard IPRs, $(S_1 \leftarrow S_2)$ and $(S_1 \leftarrow S_3)$. This is exactly the case that positive cases of IPR are $\{(S_1 \leftarrow S_2), (S_1 \leftarrow S_3)\}$, and negative cases are all sentence pairs excluding self-pointers. That is, negatives are $\{(S_2 \leftarrow S_1), (S_2 \leftarrow S_3), (S_3 \leftarrow S_1), (S_3 \leftarrow S_2 )\}$. In this case, self-pointer cases are $\{(S_1 \leftarrow S_1), (S_2 \leftarrow S_2), (S_3 \leftarrow S_3) \}$.} \footnote{For IPI, we are also able to create sentence pairs. For instance, suppose there is a parent post which contains three sentences $(S_1, S_2, S_3)$, a child post that contains two sentences $(S_4, S_5)$, and a gold standard IPI, $(S_2 \leftarrow S_5)$. The positive case of IPI is exactly $\{(S_2 \leftarrow S_5)\}$, and negative cases are all sentence pairs excluding self-pointers, that is, $\{(S_1 \leftarrow S_4), (S_1 \leftarrow S_5), (S_2 \leftarrow S_4), (S_3 \leftarrow S_4), (S_3 \leftarrow S_5) \}$.}

\subsubsection*{Baselines}
First, we employ state-of-the-art PN techniques from \cite{Potash2017} as baselines. The use of these baselines was decided because our model PCPA ({\bf Our Model}) employs pointer architectures. As the authors proposed two techniques, sequence-to-sequence model ({\bf PN with Seq2Seq}) and w/o sequence-to-sequence model ({\bf PN without Seq2Seq}), we have the two models for comparison.

\par
To analyze how a {\it non} PN model works, multi-task learning is employed to the baseline \cite{Sogaard2016} ({\bf STagBLSTM}) by \cite{eger2017}. STagBLSTM is composed of shared BiLSTM layers for subtasks, and output layers for each subtask. In \cite{eger2017}, the authors provided a BIO tagging task, however, the task is not required in our work because BiLSTM handles an input as sentence representation rather than as word representation. In this paper, we use one BiLSTM.\footnote{Though there are some variation models other than the single BiLSTM model, our preliminary experiments show a non-significant improvement.}

\par
To show end-to-end learning models are effective for AM on thread structures, we provide the following three task specific baselines. First, feature-based SVM \cite{stab2017} ({\bf SVM - T}) is introduced. $T$ indicates each subtask of the claim classifier, premise classifier, IPR classifier, and IPI classifier. In addition, random forest ({\bf RF - T}) and the logistic regression technique \cite{Peldszus2015} ({\bf Simple - T}) are also introduced. For each task specific model, BoW features the top 500 most frequent words \footnote{In fact \cite{stab2017} and employs rich features such as structural features. We only use BoW for comparison because the properties of COLLAGREE corpus substantially differ from their corpus.}.

\par
We assume that each output of PN with Seq2Seq, PN without Seq2Seq or STagBLSTM does not satisfy the constraints as a self-pointer. This is because inappropriate outputs with constraint violations of IPR and IPI by these approaches will happen, i.e., they can predict IPI out of parent and child posts. The assumption maintains the false positive (FP) of baselines, since a self-pointer which results from a chance is not counted as FP. This condition gives the baselines the advantage of precision over our models. Therefore, this assumption is convincing.

\par
The following describes our implementation details. The implementation of neural models are by Chainer \cite{chainer_learningsys2015}. The hyperparameters are the same as \cite{Potash2017} for the PN baselines and our models\footnote{Hidden input dimension size 512, hidden layer size 256 for the BiLSTMs, hidden layer size 512 for the LSTM decoder of PN without Seq2Seq, and high dropout rate of 0.9 \cite{Srivastava2014, Guido2016}. All models are trained with the Adam optimizer \cite{Kingma2014} with a mini batch size of 16.}. In the interest of time, we ran 50 epochs, and used the trained model for testing. The COLLAGREE dataset is divided into training threads and testing threads at $8:2$. In addition, we use the following hyperparameters in equation (\ref{eq:loss}): $\alpha=\beta=1/3$. However, total loss of $L^{ipr}$ and $L^{ipi}$ tends to enlarge since they have to calculate a sum of the sentence pairs. Hence, we provide a model with tuned hyperparameters $\alpha=\beta=0.15$ ({\bf Our Model - Hyp}) for comparison.

\subsection{Experimental Results}
Table \ref{tb:f1_score} summarizes the results of our models and baselines. For each model, we showed the best F1 score in the table. Due to limitations of space, we omitted recalls and some precisions. Surprisingly, all models performed as well as we expected in our dataset, in spite of low agreements (see Table \ref{tb:kappa}). Although the basis of the ACC type classifier of PCPA is the same as the PN model, our model with tuned hyperparameters is better at NonArg identification than the baseline PN models.

\par
Both of our models significantly outperform all baselines for the IPR and IPI discrimination tasks. "Our Model - Hyp" achieves F1 $+9.3\%$ in IPR identification in comparison with the best baseline PN without Seq2Seq. This is the most important result because it indicates that incorporating constrains of thread structures with the PNs makes relation classifiers	 easy to learn and discriminate. 

STagBLSTM shows lower scores in terms of both IPR and IPI identification, implying the difficulty of the use of the multi-task learning of BiLSTM. In addition, Table \ref{tb:f1_score} (Middle) also illustrates that most neural models yield better F1 scores in comparison with the task specific models. In addition, the logistic regression and RF are overfitted, despite that cross validations are employed. Thus, end-to-end learning assumes an important role for AM, even in thread structures.

\subsubsection*{Effectiveness of Separator Representation}
To demonstrate the effectiveness of the separator representations,
we conducted an experiment. In Table \ref{tb:f1_score} (Bottom), the models without the separator input representations are indicated as "w/o separator". It shows that separator representations dramatically improve scores of PN based models. This remarkable result is from the ability to learn the structural information of a thread by encoding separators in the BiLSTM.

\begin{table}[tb]
\small
\centering
\begin{tabular}{|l|c|c|}
\hline
\multicolumn{1}{|c|}{Model} & IPR - F1      & IPI - F1 \\ \hline
Our Model                   & $\pm 0.7$     & $\pm 1.8$        \\ 
PN with Seq2Seq                  & $\pm 2.3$     & $\pm 1.2$        \\
PN without Seq2Seq               & $\pm 2.7$     & $\pm 3.9$        \\ \hline
\end{tabular}
\caption{Standard deviations of F1 scores ($\%$)}
\label{tb:stdev}
\end{table}

\subsubsection*{Stability}
To analyze the stability of our models, we compare standard deviations among three selected models. Table \ref{tb:stdev} shows standard deviations for the three models. These results indicate that our model has lower standard deviations for IPR than baseline PN models. The reason for this is the size of search space: our models can effectively limit the search space based on thread structures.

\begin{table}[tb]
\small
\centering
\begin{tabular}{|l|c|c|}
\hline
\multicolumn{1}{|c|}{Model} & IPR - F1   & IPI - F1   \\ \hline
Our Model                   & *{\bf 39.6} & *{\bf 22.6} \\
Our Model with Param Share  & 36.7       & 11.9       \\ \hline
\end{tabular}
\caption{The effect of parameter sharing of the two pointer architectures.}
\label{tb:param_share}
\end{table}

\subsubsection*{Analysis for Parallel Design}
Next, we show how our models improve their performance by employing our parallel pointer architecture. Herein, we provide a new model of PCPA with a single PN ({\bf Our Model with Param Share}), which shares $v_1$, $W_1$ and $W_2$ in equation (\ref{eq:uijkl}) and $v_2$, $W_3$ and $W_4$ in equation (\ref{eq:qijkl}), respectively. Table \ref{tb:param_share} demonstrates the mean of F1 scores for our model and Our Model with Param Share. Note that the average performances are lower than the best performances in Table \ref{tb:f1_score}. The scores indicate that sharing the two pointer architecture parameters is not effective in our proposed model. We estimate this is because poor association \cite{Caruana1997} between the IPR and IPI identification tasks exists. Therefore, our approach of using two parallel pointer architectures is effective.

\begin{figure}[t]
  \begin{center}
    \begin{tabular}{c}
      \begin{minipage}{0.45\hsize}
        \begin{center}
          \includegraphics[clip, width=\linewidth]{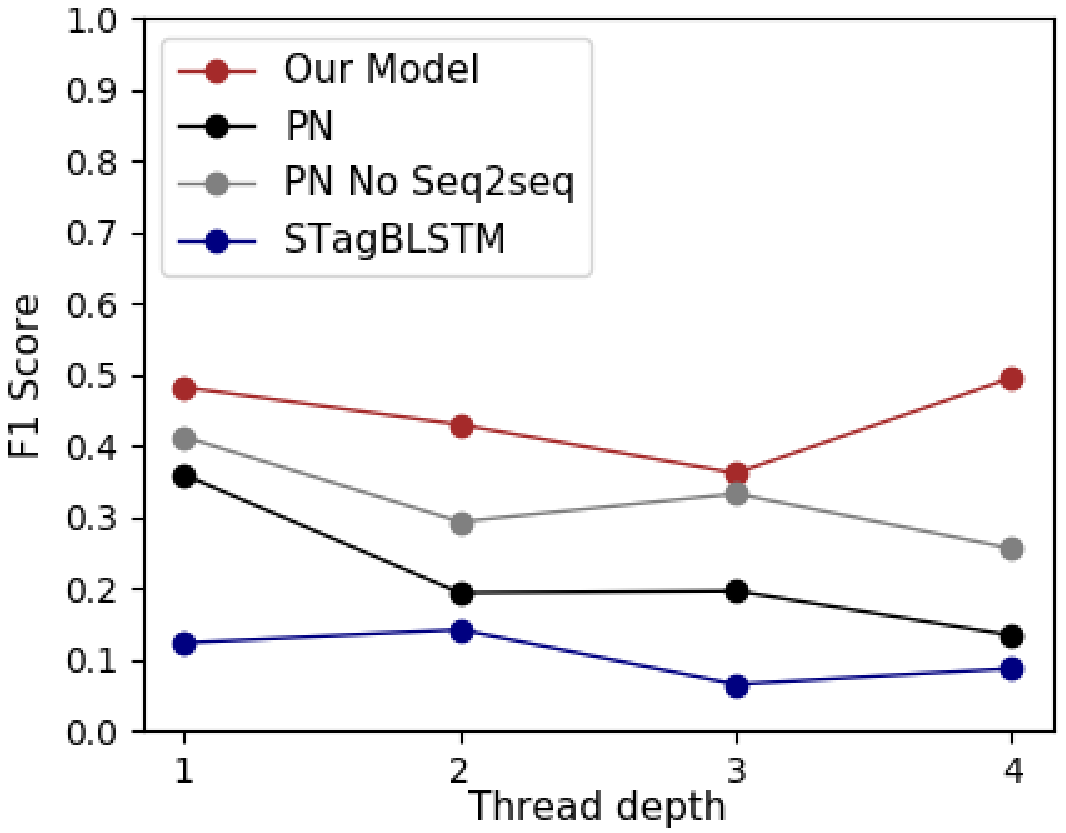}
          \subcaption{IPR}
          \label{fig:inner_f1}
        \end{center}
      \end{minipage} 
      \begin{minipage}{0.45\hsize}
        \begin{center}
          \includegraphics[clip, width=\linewidth]{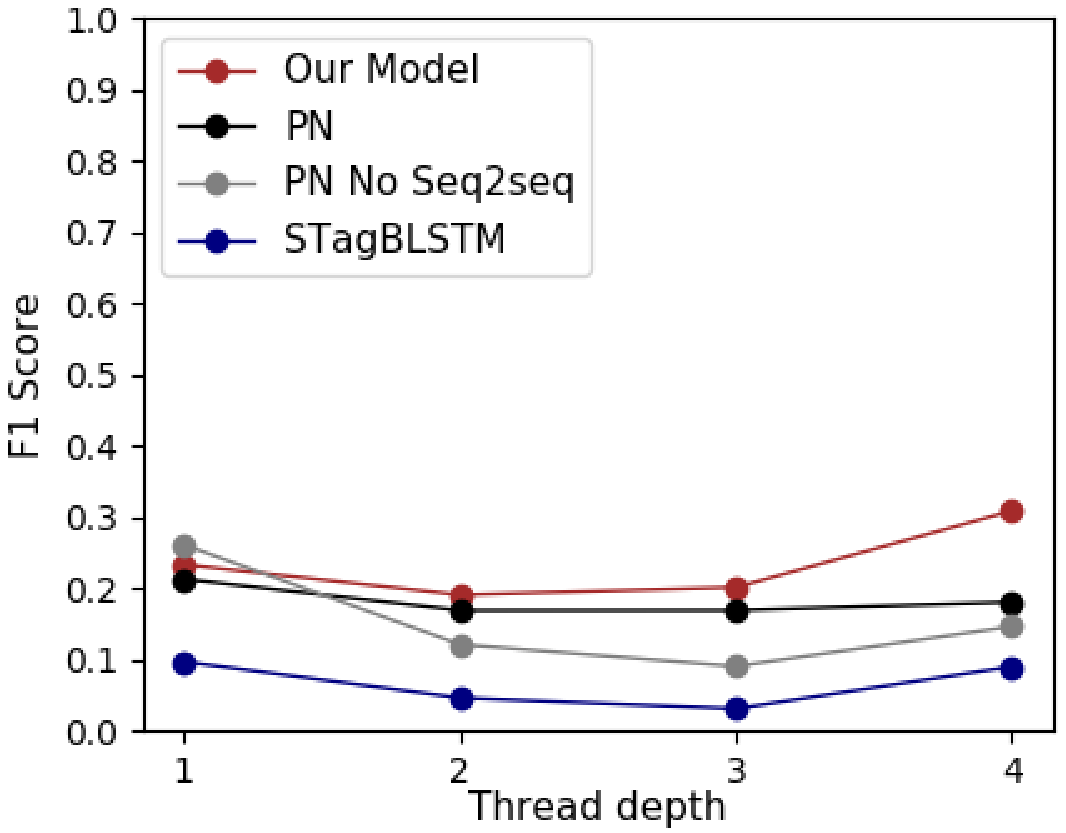}
          \subcaption{IPI}
          \label{fig:inter_f1}
        \end{center}
      \end{minipage}
    \end{tabular}
    \caption{Performances on different thread depths.}
    \label{fig:thread_depth}
  \end{center}
\end{figure}

\begin{figure}[t]
  \begin{center}
    \begin{tabular}{c}
      \begin{minipage}{0.45\hsize}
        \begin{center}
          \includegraphics[clip, width=\linewidth]{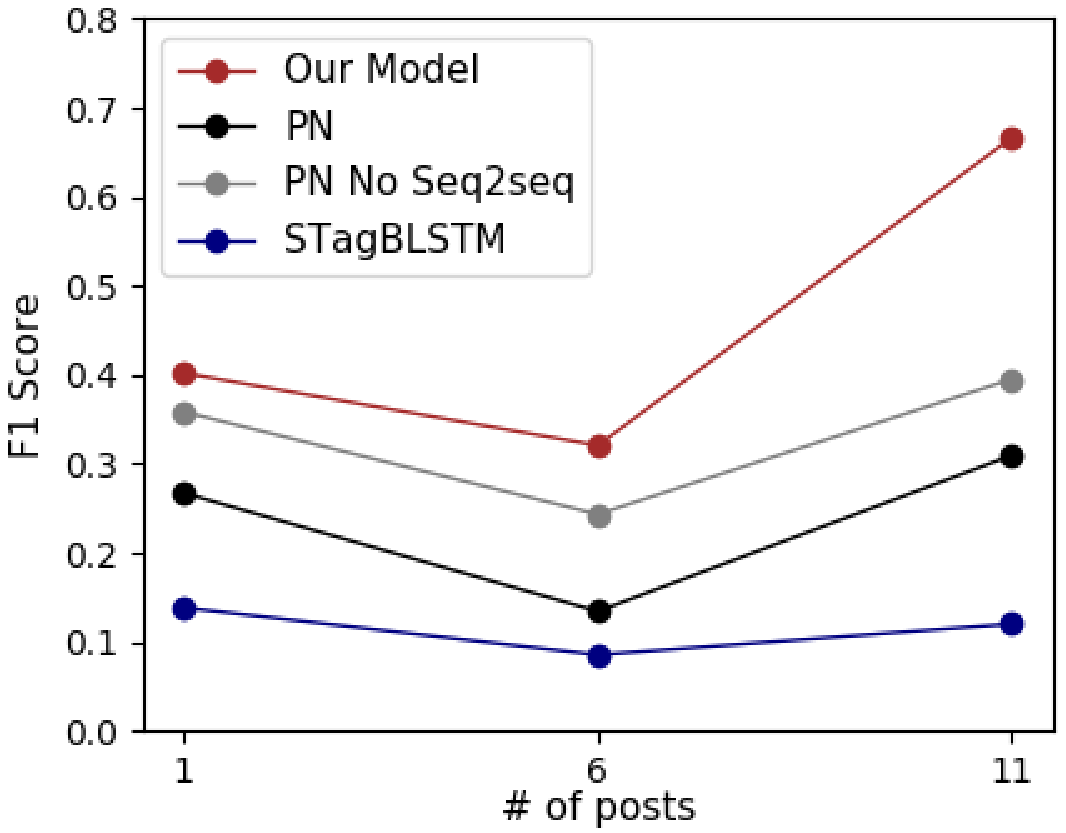}
          \subcaption{IPR}
          \label{fig:post_num_inner_f1}
        \end{center}
      \end{minipage}
      \begin{minipage}{0.45\hsize}
        \begin{center}
          \includegraphics[clip, width=\linewidth]{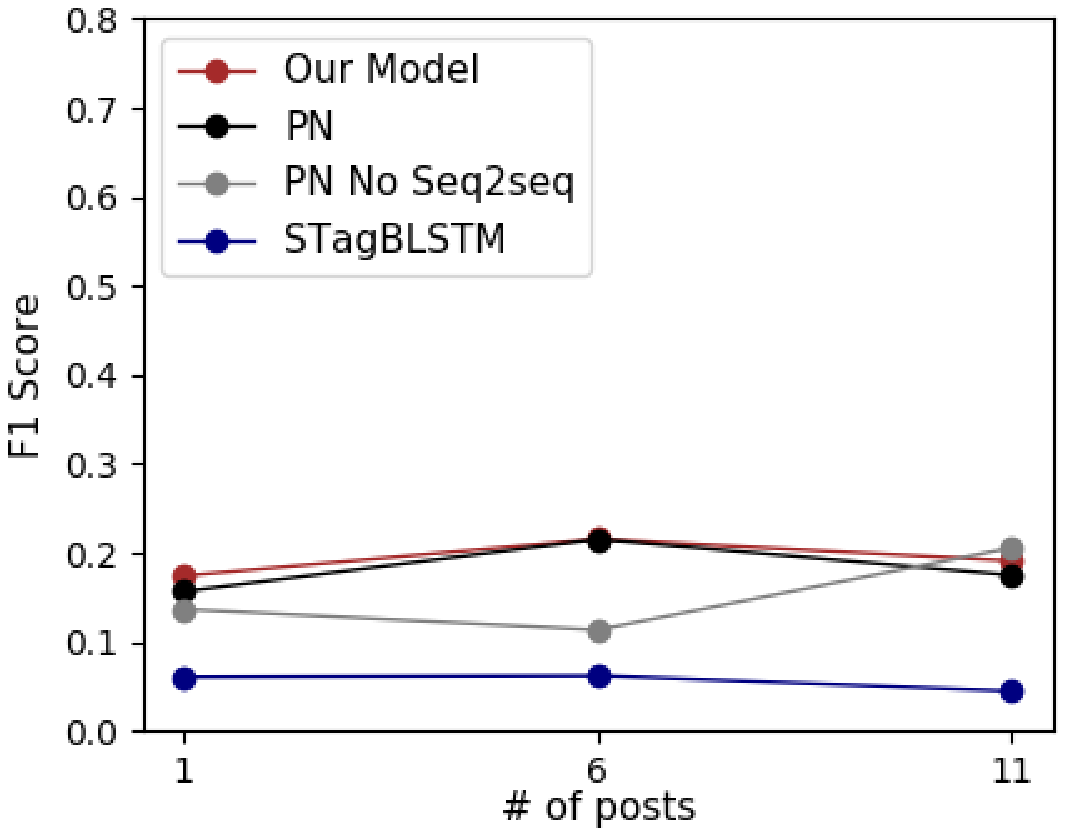}
          \subcaption{IPI}
          \label{fig:post_num_inter_f1}
        \end{center}
      \end{minipage}
    \end{tabular}
    \caption{Performances on different number of posts. When the horizontal value is 1, we test using threads which contains [1-5] posts.}
    \label{fig:post_num}
  \end{center}
\end{figure}

\subsubsection*{Performance Specialized in Threads}
We examine how our models are specialized in thread structures. Specifically, we limit the threads in test datasets by specific thresholds, and then analyze performance transitions. We conduct two experiments as the thread depth is limited (Figure \ref{fig:inner_f1} and \ref{fig:inter_f1}). While the baselines performances decrease as the thread depth increases, our model keeps its F1 score because of the separators and the search space. The separator representations for an input increase according to the thread depth, and the baseline PN models need to use wider range of hidden states in comparison with the PCPA model. In other words, our models are extremely effective, even for deeper threads.

\par
We also limit the threads that we can use in test data by the number of posts (Figure \ref{fig:post_num_inner_f1}, and \ref{fig:post_num_inter_f1}). For discriminating IPR, our model increasingly outperforms others in accordance with the number of posts. Figure \ref{fig:post_num_inter_f1} indicates that the difference between our model and baselines is minimal. This is because the number of posts does not affect the thread depth, necessarily. Most of {\it COLLAGREE}'s threads have a depth of at most 2. In other words, Figure \ref{fig:post_num_inter_f1} also implies the depth of threads affects the improvement of IPI identifications.

\section{Conclusion} \label{Conclusion}
This paper presented an end-to-end study on discussion threads for argument mining (AM). We proposed an AM scheme that is composed of micro-level inner- and inter- post scheme for a discussion thread. The annotation result shows we acquire the valid and pretty argumentative corpus. To structuralize the discourses of threads automatically, we propose a neural end-to-end AM technique. Specifically, we presented a novel technique to utilize constraints of the thread structure for pointer networks. The experimental results demonstrated that our proposed model outperformed state-of-the-art baselines in terms of relation identifications.

\par
Possible future work includes enhancing our scheme for less restricted conditions, i.e., multiple targets from one callout.

\section*{Acknowledgments}
This work was supported by CREST, JST (JPMJCR15E1), Japan and JST AIP-PRISM Grant Number JPMJCR18ZL, Japan.
We thank Takayuki Ito, Eizo Hideshima, Takanori Ito and Shun Shiramatsu for providing us with the COLLAGREE data.

\balance
\bibliography{emnlp2018}
\bibliographystyle{acl_natbib_nourl}

\end{document}